\documentclass[conference]{IEEEtran}
\IEEEoverridecommandlockouts
\usepackage{cite}
\usepackage{amsmath,amssymb,amsfonts}
\usepackage{algorithmic}
\usepackage{graphicx}
\usepackage{subfigure}
\usepackage{textcomp}
\usepackage{xcolor}
\usepackage{pgf-pie}

\def\BibTeX{{\rm B\kern-.05em{\sc i\kern-.025em b}\kern-.08em
    T\kern-.1667em\lower.7ex\hbox{E}\kern-.125emX}}
    
\begin{document}

\title{Ontology- and LLM-based Data Harmonization for Federated Learning in Healthcare\\

\thanks{}
}

\author{\IEEEauthorblockN{Natallia Kokash}
\IEEEauthorblockA{\textit{Institute of Informatics} \\
\textit{University of Amsterdam}\\
The Netherlands \\
0000-0003-3639-1245}
\and
\IEEEauthorblockN{Lei Wang}
\IEEEauthorblockA{\textit{College of Medicine} \\
\textit{The Ohio State University}\\
OH, USA  \\
Lei.Wang2@osumc.edu}
\and
\IEEEauthorblockN{Thomas H. Gillespie}
\IEEEauthorblockA{\textit{Department of Neuroscience} \\
\textit{University of California}\\
CA, USA \\
0000-0002-7509-4801}
\and
\IEEEauthorblockN{Adam Belloum}
\IEEEauthorblockA{\textit{Institute of Informatics} \\
\textit{University of Amsterdam}\\
The Netherlands \\
0000-0001-6306-6937}
\and
\IEEEauthorblockN{Paola Grosso}
\IEEEauthorblockA{\textit{Institute of Informatics} \\
\textit{University of Amsterdam}\\
The Netherlands \\
0000-0003-4600-9812}
\and
\IEEEauthorblockN{Sara Quinney}
\IEEEauthorblockA{\textit{School of Medicine} \\
\textit{Indiana University}\\
IN, USA \\
squinney@iu.edu }
\and
\IEEEauthorblockN{Lang Li}
\IEEEauthorblockA{\textit{College of Medicine} \\
\textit{The Ohio State University}\\
OH, USA \\
0000-0002-0746-1809}
\and
\IEEEauthorblockN{Bernard de Bono}
\IEEEauthorblockA{\textit{School of Medicine} \\
\textit{Indiana University}\\
IN, USA \\
0000-0003-0638-5274}
}

\maketitle

\begin{abstract}
The rise of electronic health records (EHRs) has unlocked new opportunities for medical research, but privacy regulations and data heterogeneity remain key barriers to large-scale machine learning. Federated learning (FL) enables collaborative modeling without sharing raw data, yet faces challenges in harmonizing diverse clinical datasets. This paper presents a two-step data alignment strategy integrating ontologies and large language models (LLMs) to support secure, privacy-preserving FL in healthcare, demonstrating its effectiveness in a real-world project involving semantic mapping of EHR data.
\end{abstract}

\begin{IEEEkeywords}
Federated learning, healthcare, LLM application, biomedical ontologies
\end{IEEEkeywords}

\section{Introduction}\label{sect:intro}

The rapid digitalization of healthcare has led to the proliferation of electronic health records (EHRs), offering unprecedented opportunities for data-driven medical research and clinical decision-making~\cite{meystre2008extracting,Gibson2021}. However, leveraging this data at scale remains challenging due to stringent privacy regulations, security risks, and ethical concerns associated with centralized data storage. Traditional machine learning (ML) approaches rely on the aggregation of patient data into centralized repositories, making them vulnerable to massive data breaches and non-compliance with regulations such as the Health Insurance Portability and Accountability Act (HIPAA) in the United States and the General Data Protection Regulation (GDPR) in Europe~\cite{HIPAA,GDPR2016}. To address these concerns, federated learning (FL) has emerged as a promising paradigm for collaborative model training without the need for direct data sharing~\cite{Kairouz2021}.

Federated learning enables multiple healthcare institutions to run ML models locally on their private datasets while only sharing model updates instead of raw patient data~\cite{Li2020}. This decentralized approach significantly reduces the risk of data leaks and helps maintain compliance with privacy regulations. Despite these advantages, FL in healthcare presents several challenges, particularly in terms of data heterogeneity, security vulnerabilities, and computational overhead~\cite{Rieke2020,Bhanbhro2024}. One major obstacle is the alignment and harmonization of heterogeneous EHR formats, which can vary significantly across institutions due to differences in clinical terminologies, data collection standards, and infrastructure~\cite{Haendel2018}.
Data harmonization is the practice of ``reconciling'' various types, levels and sources of data in formats that are compatible and comparable, and thus useful for better decision-making~\cite{adhikari2020cohort,abbasizanjani2023harmonising}.

Data harmonization often relies on probabilistic and/or ML-based entity resolution techniques~\cite{doan2012principles}. Schema matching~\cite{Hellenberg2025PRISMAAP} automates the identification of correspondences between fields in different datasets, such as aligning "DOB" in one database with "DateOfBirth" in another. Modern tools leverage natural language processing (NLP) and ontology-based reasoning to improve accuracy~\cite{zhang2013unsupervised}. Type conversion ensures consistent representation of data types, such as converting blood pressure values stored as strings into standardized numeric formats or translating medication codes between vocabularies like RxNorm~\cite{Nelson2011} and SNOMED CT~\cite{Donnelly2006}. In healthcare, these automated techniques support critical applications like patient cohort identification~\cite{adhikari2020cohort}, population health monitoring, and real-time clinical decision support, reducing manual curation and improving data quality for ML pipelines and interoperable health information systems.

Large language models (LLMs), trained on vast corpora of biomedical literature and structured clinical data, have demonstrated strong capabilities in natural language understanding and information extraction~\cite{JAHAN2024108189}. They can be leveraged to standardize disparate EHRs, align ontologies, and mitigate discrepancies in medical coding practices across different hospitals and research centers~\cite{wang2018clinical}. However, ensuring the trustworthiness, bias mitigation, and interpretability of LLMs in clinical applications remains a critical research frontier~\cite{ALBAHRI2023156,Chen2024}. 

This paper explores the intersection of FL and healthcare, focusing on data harmonization strategies within privacy-preserving data access environments. With attention to security and regulatory compliance challenges, we are working on the integration of LLM-based functionality to a programmable FL framework to enable healthcare data alignment. We show how our two-step ontology- and LLM-based data alignment strategy was instrumental in the mapping of healthcare data for a real-world project. In the first step, the converter generated matching candidates using (a) vector-space embeddings~\cite{mikolov2013efficient} and/or (b) ontology-based converter matching. In the second step, an LLM was used to accept or reject the matching pairs.      

The paper is structured as follows. In Section~\ref{sect:FL} we introduce the FL framework that we intend to empower with LLM support to improve the experience of research scientists designing workflows. In this work, we focus on the integration of LLMs to functions that perform data harmonization at domain nodes. Section~\ref{sect:onto} discusses the problem of healthcare data harmonization focusing on disambiguation via alignment with biomedical ontologies. We propose an LLM-empowered pipeline to automatically convert natural language annotations to the corresponding ontology codes. Section~\ref{sect:mprint} introduces a collaborative real-world project in healthcare as a use case where the pipeline was applied to handle semantic heterogeneity of EHRs data. In section~\ref{sect:related-work}, we overview related work. Section~\ref{sect:conclusion} concludes our work and outlines future steps.

\section{Federated learning frameworks}\label{sect:FL}

FL applications face issues such as diversity in data types, model complexity, privacy concerns, and the need for efficient resource distribution. The research communities have been working on minimizing the effort by designing dedicated frameworks, reusable architectural patterns, and domain specific languages to orchestrate workflows and express security and privacy policies. Among such frameworks are Vantage6~\cite{Moncada-Torres2021} and Brane~\cite{9582292} which jointly provide an infrastructure to design, deploy, and run federated workflows. 

\subsection{Vantage6}
Vantage6~\cite{Moncada-Torres2021} allows researchers to perform ML operations on client's data located at worker (computing) nodes. The process is orchestrated by a server (central node). A researcher can submit a task to the server with an algorithm and input parameters. The algorithm is first implemented using Vantage6 tools and built into a Docker image~\cite{docker}. After a task is submitted to the server, the server sends the task information to a computing node. A computing node automatically detects the server, gets the task information, and executes the algorithm on local data. The intermediate results are sent back to the server for aggregation, and the iterative process of FL repeats to update the global model. The final result is sent back to the researcher when the computation is complete. 

\begin{figure}[htbp]
    \centering
    \includegraphics[width=0.45\textwidth]{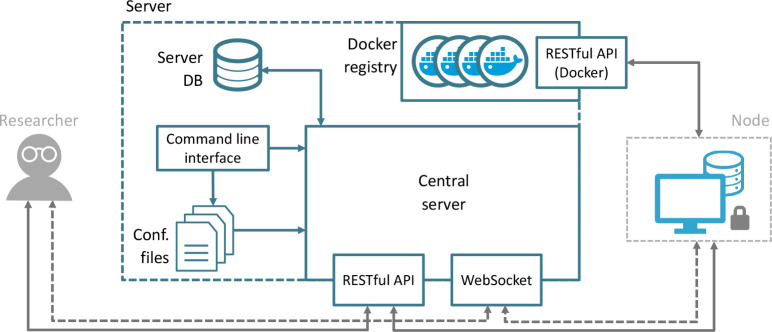}
    \caption{VANTAGE6~\cite{Moncada-Torres2021} The server loads its configuration parameters and exposes its RESTful API Nodes.}
    \label{fig:vantage6}
\end{figure}

\subsection{Brane}

Brane is a programmable framework for secure data exchange and scientific workflow orchestration. The primary purpose of Brane is to support efficient and secure data exchange among research organizations~\cite{9582292}. Brane utilizes containerization to encapsulate functionalities as portable building blocks. Through programmability, application orchestration can be expressed using an intuitive domain-specific language or user-friendly interactive notebooks. End users with limited programming experience are empowered to compose workflows without the need to deal with the underlying technical details. 

\begin{figure}[htbp]
    \centering
    \includegraphics[width=0.49\textwidth]{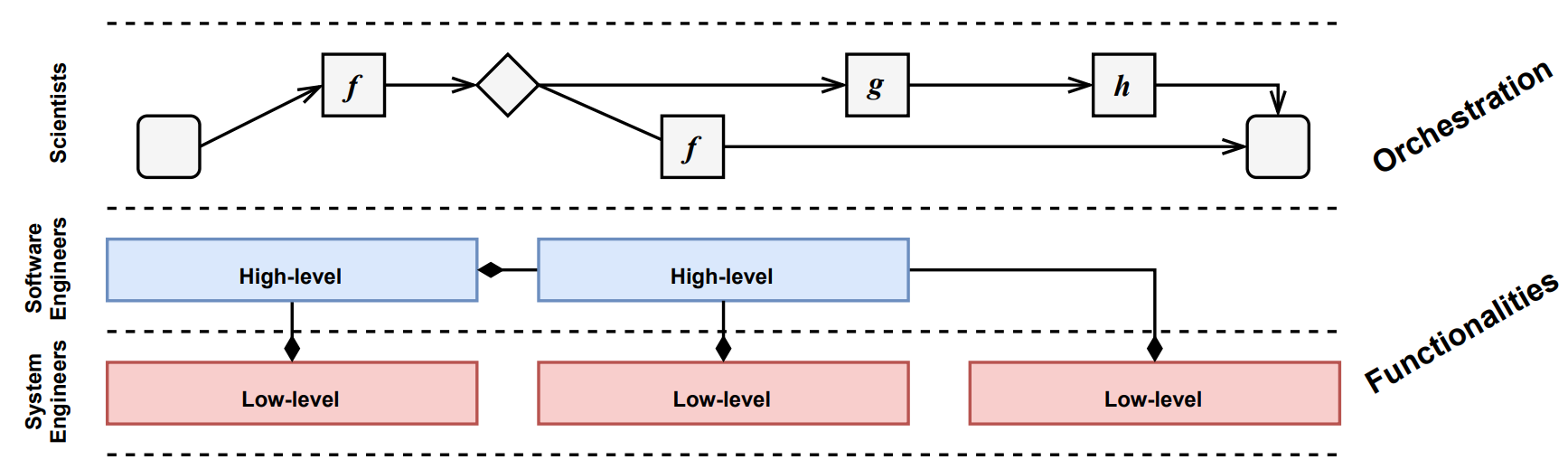}
    \caption{Brane's approach to the distributed workflow implementation via separation of user roles~\cite{9582292}.}
    \label{fig:brane}
\end{figure}

A key principle of Brane's design is the clear separation of concerns based on specialized user roles, as shown in Figure~\ref{fig:brane}:
(i) \emph{domain scientists} focus on data analysis without managing execution details, (ii) \emph{software engineers} develop and optimize data processing workflows, (iii) \emph{systems engineers} maintain the infrastructure and ensure system efficiency.

The only requirement for remote resources is to be able to install and run containers.
Brane supports Docker~\cite{docker}, Singularity~\cite{Kurtzer2017}, and Kubernetes clusters~\cite{burns2016borg}. The runtime system can automatically convert packages (Open Container Initiative (OCI) images~\cite{oci}) to the appropriate container image format. By default, direct access to resources is assumed; when this is not possible (e.g., not permitted by participating organizations or regulatory policies), an optional indirection layer is enabled.

\subsection{EPI platform}
Brane is a key component of the Enabling Personalized Interventions (EPI) framework~\cite{alsayedkassem_et_al:OASIcs.Commit2Data.2}. The EPI framework provides a secure, distributed data platform that supports personalized health insights through analytics and decision support tools. Its main features are:
\begin{itemize}
    \item Allowing analysts to process data across multiple organizations without dealing with technical complexities.
    \item Enforcing user-defined data policies during all stages of data processing.
\end{itemize}

Various extensions were added to Brane to provide a seamless experience for data scientists while enforcing strict data protection measures. The framework automates the setup of the underlying infrastructure while considering the different requirements communicated by its components.

\begin{figure}[htbp]
    \centering
    \includegraphics[width=0.49\textwidth]{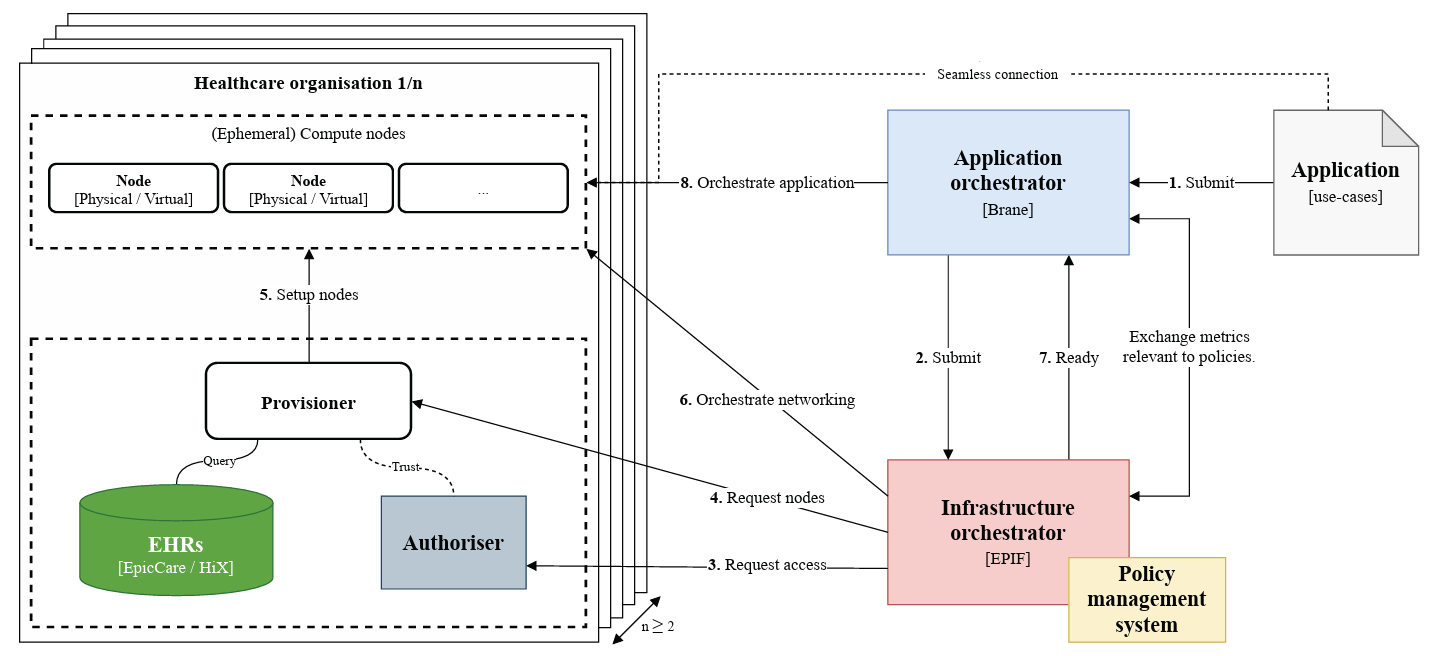}
    \caption{EPI Framework~\cite{kassem2021epi}.}
    \label{fig:epi}
\end{figure}

Figure~\ref{fig:epi} shows the main components of the framework, which are:
\begin{itemize}
    \item the \emph{orchestrators} (both at the application level and infrastructural level);
    \item the \emph{policy management system};  
    \item the components required to be present at the participating institutions: the \emph{resource provisional} and the \emph{authorizers}.
\end{itemize}

At the network level, the EPI platform enforces security and low-level policies to protect data sharing. 

\subsection{Discussion}
In collaboration with health organizations (St.Antonius Ziekenhuis, UMC Utrecht, and Princes Maxima Centrum), a proof-of-concept collaborative network has been deployed to process patient data across the two hospitals and the Dutch supercomputer center SURF~\cite{alsayedkassem_et_al:OASIcs.Commit2Data.2}. This network is a successful application of the EPI framework for collaborative research on privacy-sensitive data in the healthcare domain. 

The EPI project focused on fast and secure computations across healthcare institutions; it prioritized usability, data privacy, and security. However, it has been applied within a closed consortium of organizations that agreed to participate in a joint FL project. The partners: 
\begin{itemize}
    \item agreed on a common research project a priori;
    \item exchanged information, discussed data content, access strategies, and agreed on learning workflows;
    \item worked together to prepare and align data for FL;
    \item discussed and agreed on compliance beforehand;
    \item trusted each other to act ethically in protecting data.
\end{itemize}

Architectural FL solutions and workflow management frameworks often leave the problem of data alignment to their users, as these problems are application specific. Domain experts are not always able to meet the technical requirements imposed by the FL platforms. This hinders their practical applications. 

Regulatory requirements such as GDPR and HIPAA are not the major concern within closed consortia. For example, the GDPR's ``right to be forgotten'' is often not the part of the automated workflow, i.e., participating organizations cannot request for their data, metadata, user profile, etc. to be removed or rectified via the platform's API requests. 

The FL scope and architectural pattern~\cite{lo2022architectural,liu2024agent}, data content, schemata, and organizational conventions are known beforehand. These aspects make the FL workflow design and orchestration somewhat easier than in an ``open-ended'' scenario where there is no prior knowledge about the partners and their data. However, the FL solutions designed under these conditions are hardly ever reusable. The workflows are hard to reproduce even for identical projects because the FL methods are designed to work with apriori known data (types, formats, distributions, dimensions, and so on).    

\subsection{Towards an Open-FL platform}

A unique selling point of the Brane's framework lies in its flexibility. Unlike many specialized collaborative and FL platforms in healthcare~\cite{NASARIAN2024102412,Topaloglu2018,PMID:34040261}, it provides a technology for anyone interested in secure private data access to develop their own solution. In this respect, and provided that a good support infrastructure becomes available, this tool has potential similar to enterprise low-code application platforms~\cite{gartner_lcap_reviews} (for which the global market is experiencing significant growth - valued USD 24.83 billion in 2023 and is projected to reach USD 101.68 billion by 2030~\cite{grandview2024lowcode}).   

We are working on the adaptation of the Brane/EPI frameworks to serve open consortia of (healthcare) organizations interested in on-demand FL networks. In our vision, any scientist interested in the deployment of an FL network to answer a research question, (i) design a high-level workflow and, (ii) in collaboration with software and system engineers, deploys a project server that allows any interested organization with potentially relevant data to (iii) join the call by downloading a pre-configured container, through which the scientist (iv) pushes ML algorithm images, executes them on local data, and (v) receives the results back to the server for aggregation. 

The combination of Vantage6, Brane and EPI tools are able to provide technical solutions for multiple issues relevant to this vision. Although Brane is not a dedicated FL platform, its programmable nature and universality make it suitable for FL workflow implementation. L.Liu \cite{liu2021federated} in her thesis showed how the Vantage6 FL algorithm images can be deployed and executed on a Brane network. Notably, this work mentioned the lack of data converters (even simple data type conversions) as the main obstacle to the deployment of FL workflows. 

The main difficulty in realizing this vision is the lack of ready-to-use resources and a community support network to keep the FL workflow process design ``low-code''. Dhooper~\cite{Dhooper2020} discusses the advantages of the agnostic approach to ML. In particular, a data-agnostic approach signifies the ability of a learning system to process the data collected from heterogeneous data sources. The ML model should be designed in such a way that it can process unstructured data seamlessly as it processes structured data. A collection of data-agnostic ML methods available for the use within Brane containers would significantly increase the prospects of the framework's application for FL research. While, as many  student projects showed, the integration of Python-based ML libraries (e.g., PyTorch~\cite{yamwd-pytorch-interface-2024}) within Brane is rather straightforward, regulatory compliance and data harmonization are two challenging aspects of a FL process that research scientists are left to implement on the application level.  

The proven success of large language models (LLMs) in processing unstructured data, requests in natural language, code generation, summarization, transformation, and data mapping makes them a promising tool for bridging the gaps within FL workflows. Hence, we aim at integrating LLM-based assistants to the Brane/EPI framework to:
\begin{itemize}
    \item simplify compliance policy translation to the specification formats supported by the framework (eFlint~\cite{10.1145/3425898.3426958}, Datalog~\cite{esterhuyse2025justact});
    \item enable data harmonization pipelines to overcome structural and semantic heterogeneity of federated data. 
\end{itemize}
In this paper, we focus on the second aspect of this roadmap. In particular, we integrate an LLM to provide a data-agnostic conversion function that aligns patient EHRs with standardized biomedical vocabularies. 

\section{Aligning biomedical data via ontologies}\label{sect:onto}

Biomedical ontologies such as SNOMED CT~\cite{Donnelly2006}, ICD-10~\cite{WHO2004}, MONDO~\cite{Mungall2022}, and HPO~\cite{Kohler2021} were created to standardize the representation of medical concepts, enabling accurate communication, data integration, and interoperability across healthcare and research domains. SNOMED CT, used in over 50 countries, supports EHRs and clinical decision-making~\cite{LEE201387}. ICD-10, maintained by the World Health Organization (WHO), is the global standard for disease classification, used in over 150 countries for epidemiology, billing, and public health monitoring. 
HPO standardizes descriptions of human phenotypes, widely adopted in genomic diagnostics and rare disease research. These ontologies are crucial for improving patient care, medical research, and health data analytics worldwide.
MONDO integrates multiple disease ontologies to unify rare disease research across organizations like Orphanet, OMIM, and ClinGen. The Orphanet Rare Disease ontology (ORDO) is jointly developed by Orphanet and the EBI to provide a structured vocabulary for rare diseases.

The biomedical domain encompasses an immense variety of terminologies to represent diseases, diagnoses, treatments, laboratory findings, and clinical outcomes. 
Table~\ref{tab:onto} summarizes the purpose, key features and use cases of biomedical ontologies relevant to our work. The aforementioned ontologies and resources provide structured vocabularies that help researchers and clinicians communicate and exchange medical data. However, these ontologies differ in their focus, level of granularity, and intended applications. Some are used for clinical documentation (SNOMED, ICD), some for genetic research (HPO, ClinGen, OMIM), others for pharmacology and treatment classification (RxNorm~\cite{Nelson2011}, ATC~\cite{WHO2021}, MedDRA~\cite{Brown1999}). Furthermore, medical knowledge is constantly evolving, requiring frequent updates to these ontologies. This results in multiple versions and implementations across institutions and countries, making interoperability a significant challenge. Even widely used ontologies such as SNOMED CT and ICD-10 have multiple regional implementations and undergo frequent updates~\cite{LEE201387}. 

Efforts such as the OHDSI (Observational Health Data Sciences and Informatics) Common Data Model, the LOINC-SNOMED harmonization initiative, and the UMLS Metathesaurus are crucial in ensuring that FL models trained on distributed datasets can produce reliable, interpretable, and generalizable results. However, full automation of this process remains an open research challenge, requiring advances in ontology alignment, machine learning-driven entity resolution, and human-expert validation.

\begin{table*}
\caption{Biomedical ontologies}
\begin{center}
    \begin{tabular}{|p{1.5cm}|p{5cm}|p{5cm}|p{5cm}|}
        \hline
        \textbf{Name} & \textbf{Purpose} & \textbf{Key Features} & \textbf{Use Cases} \\
        \hline
        SNOMED CT & Standardized clinical terminology for electronic health records & Provides hierarchical relationships and standardized codes for diseases, symptoms, and procedures & Clinical documentation, decision support, interoperability in healthcare IT systems \\
        \hline
        ICD-10 & Global classification of diseases, maintained by WHO & Provides alphanumeric codes for diseases and health conditions & Public health monitoring, insurance claims, medical record standardization \\
        \hline
        MONDO & Unified disease ontology integrating multiple sources & Harmonizes data from Orphanet, OMIM, and DOID & Research in rare diseases, precision medicine, biomedical data integration \\
        \hline
        HPO & Standardized vocabulary for human disease phenotypes & Describes phenotypic abnormalities in a hierarchical structure & Clinical genomics, rare disease research, computational phenotyping \\
        \hline
        ORDO & Orphanet Rare Disease Ontology, European database for rare diseases and orphan drugs & Disease and gene information & Used in rare disease research, clinical guidelines, regulatory agencies \\
        \hline
        OMIM & Online Mendelian Inheritance in Man, a catalog of human genes and genetic disorders & Provides gene-disease relationships, clinical descriptions, inheritance patterns & Used in clinical genetics, genomic research, and precision medicine \\
        \hline
        ClinGen & Clinical Genome Resource, NIH database of clinically relevant genetic variants & Assists in variant classification for genetic diagnostics & Supports clinical diagnostics, personalized medicine, and regulatory decisions \\
         \hline
        EBI & Bioinformatics resource for genomics, proteomics & Provides large-scale biological data & Used in genomics, pharmacology, and data integration \\
        \hline
        MedGen & Clinical genetics database & Organizes human genetic conditions & Used in genetic counseling, rare disease research \\
        \hline
        MeSH & Standardized biomedical terminology & Indexes PubMed, MEDLINE, and biomedical databases & Used in literature indexing and biomedical research \\
        \hline
        UMLS & Unified medical terminology system & Maps multiple vocabularies for interoperability & Used in clinical decision support, EHRs, NLP \\
        \hline
        OBO & Collection of interoperable ontologies & Supports semantic consistency in biological data & Used in bioinformatics and AI-driven research \\
        \hline
        DOID & Standardized disease classification & Links genetic and environmental factors in diseases & Used in genomics, precision medicine, and disease annotation \\
        \hline
        NCIT & Oncology-specific vocabulary & Standardizes cancer terminology & Used in clinical trials, cancer research, and informatics \\
         \hline
        RxNorm & Standardized drug terminology maintained by the U.S. National Library of Medicine & Provides normalized names and unique identifiers for clinical drugs, linking various national drug terminologies & Used in electronic health records (EHRs), clinical decision support, and pharmacy systems \\
        \hline
        ATC & Anatomical Therapeutic Chemical Classification System, classifies  drugs based on mechanism of action and therapeutic use & Hierarchical classification of drugs into five levels, covering active ingredients and therapeutic groups & Used in pharmacoepidemiology, drug regulation, and healthcare analytics \\
        \hline
        MedDRA & Medical Dictionary for Regulatory Activities, terminology for medical conditions, adverse events, and drug safety monitoring & Maintains hierarchical coding for diseases, symptoms, and adverse drug reactions & Used in pharmacovigilance, clinical trials, and regulatory reporting \\
        \hline
    \end{tabular}
    \label{tab:onto}
\end{center}
\end{table*}

\begin{figure*}[htbp]
    \centering
    \includegraphics[width=0.85\textwidth]{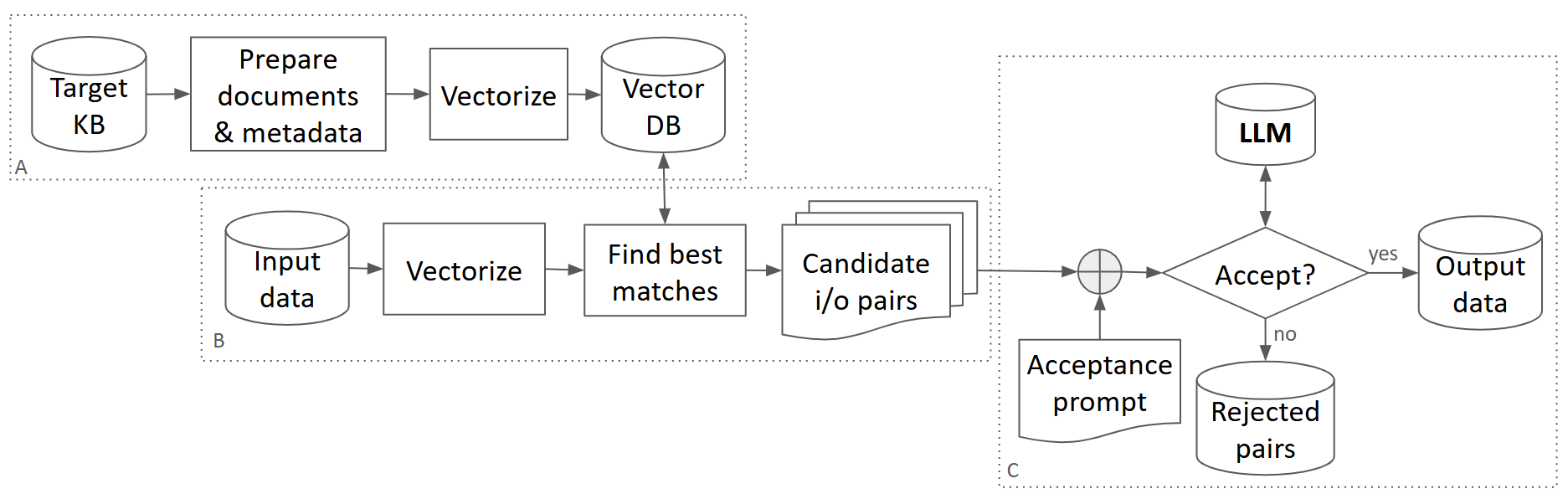}
    \caption{LLM-based pipeline to align data with target vocabulary. (A) Prepare target mapping space; (B) Find best matching targets for input data; (C) Define an acceptance criteria for LLM and evaluate best matching pairs.}
    \label{fig:llm-general}
\end{figure*}

Annotating clinical observations from EHRs with ontology terms is useful for selection of patient cohorts and creation of federated datasets for ML training and evaluation on images or laboratory tests of patients with certain diseases or pathologies. Figure~\ref{fig:llm-general} presents a generic LLM-based conversion process to map unannotated data to a target ontology terminology. The same process can be used to align the data with alternative annotations. The first step consists of (A) enabling Retrieval Augmented Generation (RAG)~\cite{10.5555/3495724.3496517} on the target vocabulary space to (B) find the best matching pairs of input data with the standardized terms. The second step consists of (C) formulating the acceptance criteria and asking an LLM to evaluate each generated matching pair, providing the criteria and the pairs in the request prompt.     

\section{Data alignment for drug reporting use case}\label{sect:mprint}
The Maternal and Pediatric Precision in Therapeutics (MPRINT) hub aggregates, presents, and expands the available knowledge, tools, and expertise in maternal and pediatric therapeutics to the broader research, regulatory science, and drug development communities. It conducts therapeutics-focused research in obstetrics, lactation, and pediatrics while enhancing inclusion of people with disabilities~\cite{mprint2023}.

The MPRINT working group processes data from multiple healthcare organizations. Relevant data encompass a wide range of information including patient demographics, diagnoses, medications, procedures, treatment history, laboratory tests, and diagnostic images. In this section, we present our evaluation of a combination of ontology and LLM-based pipelines to align textual data for an FL study within the MPRINT initiative. This particular MPRINT study is focused on a drug reporting use case that aims to establish the relationship between exposure to certain medications or chemical substances during pregnancy and their effect on pregnancy, postpartum, and/or newborn health.         

\subsection{Unannotated dataset}
The first dataset, provided by Kids First DRC - Pediatric Cancer and Rare Disease Care~\cite{kidsfirstdrc} - included 512 clinical records with pregnancy characteristic/risk factors, exposure to drugs or chemicals, and outcomes of such exposure on the pregnancy, postpartum, and neonatal conditions. The information is provided in a table with 3 fields without ontological annotations. The goal of the data harmonization pipeline was to map the textual descriptions from these set to the matching ontology terms. For brevity, in this paper we focus on mapping pregnancy outcome descriptions to the MONDO and/or HPO ontologies. 

MONDO and HPO are complementary: MONDO standardizes disease definitions, while HPO defines phenotypic abnormalities in human diseases. They are linked through annotations — MONDO diseases are often associated with specific HPO terms to describe their characteristic clinical features. Therefore, using labels from both ontologies to map the EHR records helps to improve mapping recall.

In an \textit{ontological annotation task}, where concepts from an ontology are assigned to data (e.g., EHR text or images), \textit{precision} is the proportion of predicted ontology terms that are correct, and \textit{recall} is the proportion of relevant ontology terms that were successfully predicted~\cite{manning2008introduction}.
Precision reflects how many of the assigned annotations are relevant and recall measures how many of all relevant annotations were assigned.

\begin{figure*}[htbp]
    \centering
    \includegraphics[width=0.85\textwidth]{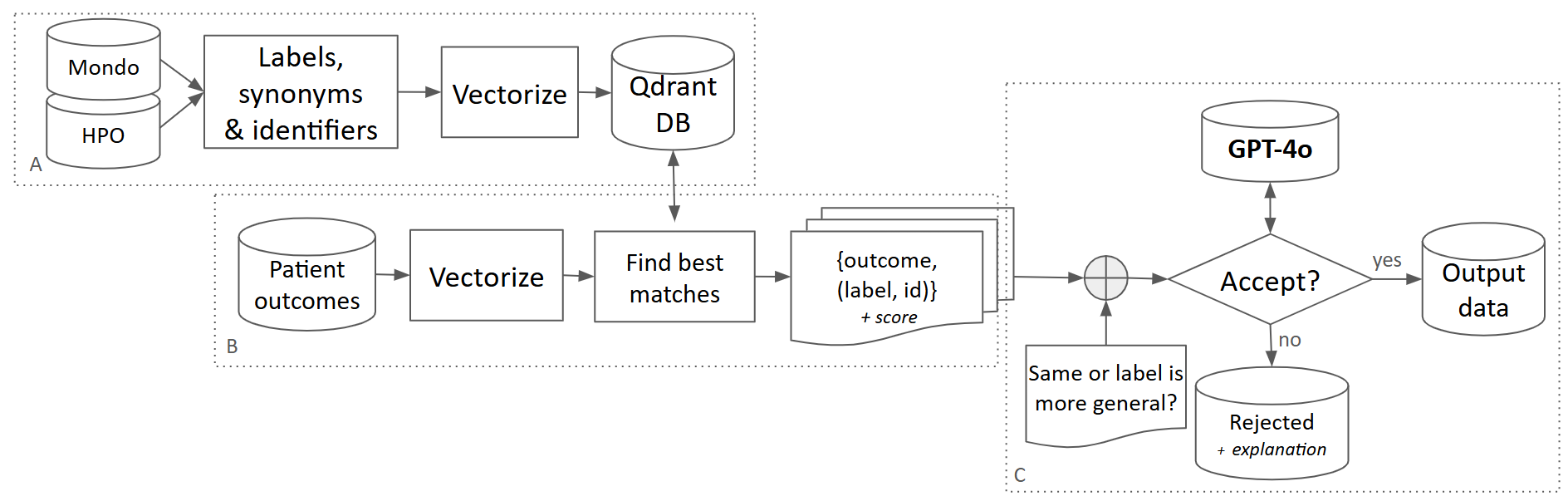}
    \caption{LLM-based pipeline to annotate patient outcomes with MONDO and/or HPO ontology terms. (A) Extract labels and synonyms from target databases, retain ontology identifiers as metadata; (B) Find best matching labels for each outcome and form pairs; (C) Ask LLM to accept pairs with the same or more general target outcomes (diseases).}
    \label{fig:llm-instance}
\end{figure*}

Figure~\ref{fig:llm-instance} shows a variant of the generic matching pipeline to translate patient outcomes from this dataset to the corresponding MONDO and HPO ontology terms. In the first step, we extract labels and synonyms from MONDO and HPO ontologies, retaining the corresponding ontology identifiers as metadata. We then create vector embeddings for these documents and store them in a Qdrant vector database cluster~\cite{qdrant2023}. For each row in the dataset, we embed the observed outcomes and use as a query to retrieve (up to) 3 most relevant MONDO/HPO disease terms and/or hereditary conditions. 
In the second step, we request the LLM (ChatGPT-4o) to decide whether the retrieved matching pairs, 1401 in total, satisfy the acceptance criteria using the following prompt:

\begin{quote}
    \small{\emph{
Given two short descriptions, decide whether they refer to the same disease or medical condition. 
If the second description is more narrow or specific, choose "No" as an answer.
If the second description is broader or more generic, choose "Yes" as an answer.
Start your answer from "Y" for "yes" or "N" for "no" and provide a concise justification, no more than 30 words, why you came to this conclusion. 
}}
\end{quote}

\begin{figure}[htbp]
    \centering
    \subfigure[Initial evaluation]{
    \label{fig:pie_chart1}
    \begin{tikzpicture}
        \pie[
            text=legend,
            sum=1401,   
            hide number,
            radius=1,    
            color={green, yellow, orange, red} 
        ]{
            352/yes-yes (352),
            933/no-no (933),
            98/no-yes (98),            
            18/yes-no (18)
        }
        \pie[sum=1401, radius=1, color={green, yellow, orange}]{352/,933/,98/}
    \end{tikzpicture}
    }
    \subfigure[Revised evaluation]{\begin{tikzpicture}
    \label{fig:pie_chart2}
        \pie[
            text=legend, 
            sum=1401,   
            hide number,
            radius=1,    
            color={green, yellow, orange, red} 
        ]{
            352/yes-yes (352),
            944/no-no (944),
            98/no-yes (98),            
            7/yes-no (7) 
        }     
        \pie[
            sum=1401, radius=1, color={green, yellow, orange} 
        ]{
            352/,944/,98/
        }
    \end{tikzpicture}}
    \caption{Number of data record mappings approved and rejected by a human expert vs LLM. }
    \label{fig:eval}
\end{figure}
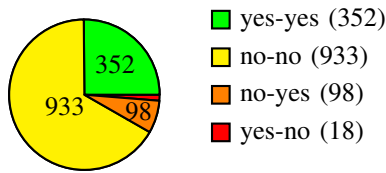
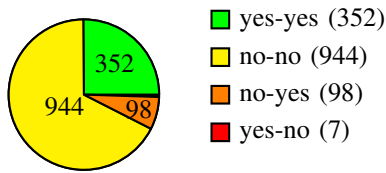

To evaluate the precision of the LLM's decisions on the equivalence of the conditions in the queries, we asked a human expert (MD), to evaluate whether diseases in the matching pairs refer to equivalent or different conditions in the given context. We then compared the accepted and rejected pairs from the human expert and the LLM. The results are summarized in Figure~\ref{fig:pie_chart1}. The decisions of the human expert and the LLM coincided in 1285 cases (92\%). In 18 cases, the human expert's decision was positive while the LLM rejected these pairs, and in 98 cases, the human expert rejected the mappings while the LLM approved them. Among these cases, 57 wrongly approved by LLM mappings referred to related outcomes, but the target description was more restrictive than the input; only 27 approved pairs referred to different diseases. 

The human expert was asked to review the decisions for pairs in which his assessment disagreed with the decision by the LLM. We clarified the requirements for the assessment of related outcomes similarly to the LLM prompt, asking to accept the mapping only if the target mapping is the same or more generic. The human expert retracted 11 out of 18 initially accepted mappings that he considered acceptable for the study context but that formally did not match the aforementioned relation. The revised results are shown in Figure~\ref{fig:pie_chart2}. 

Table~\ref{tab:llm_error} and Table~\ref{tab:human_error} give examples of EHR observations and suggested MONDO/HPO labels misjudged by the human expert and the LLM, respectively. It is easy to see why this task was challenging: the records were very concise, involved numbers, signs, and abbreviations.      

\begin{table*}
\caption{Examples of mismatched conditions by LLM}
\begin{center}
    \begin{tabular}{|p{3.4cm}|p{3.4cm}|p{10cm}|}
        \hline
        \textbf{Patient record} & \textbf{Suggested mapping} & \textbf{Explanation} \\
        \hline
        Trimester MOUD initiated - 1st &  Late first trimester onset & The first description mentions MOUD (Medications for Opioid Use Disorder), the second is about an unspecified condition observed toward the end of the first trimester. \\
        \hline
        Intraventricular hemorrhage $>$ grade 2  &  Grade II preterm intraventricular haemorrhage & The first description refers to the grade above 2, the second to the grade 2. \\
       \hline
        Congenital laryngomalacia & Acquired laryngomalacia & Congenital condition is present at birth, acquired develops after birth. \\
        \hline
        APGAR score minute 1  &  10-minute APGAR score of 1 & The first description refers to the test performed at 1st minute after birth, the second at 10th minute.\\
        \hline
        Continued vomiting hours 1-4  &  Frequent vomiting & Vomiting persists over time without significant breaks as opposed to vomiting that happens many times.\\
        \hline
        Neonatal death  &  Neonatal lethal & The first description refers to the observed (confirmed) outcome, the second - to a severe condition likely causing death.\\
        \hline
        Respiratory & Respiratory infections & Respiratory problems are not necessarily caused by infections, the second description is more restrictive.\\
        \hline
        Loose stools  &  Frequent stools & The first description refers to  consistency, the second to frequency (it may be normal in consistency but happens more often than normal)\\
        \hline
        Pharmacologic treatment of Nas  &  Nasal congestion & NAS (Neonatal Abstinence Syndrome) refers to the use of medications to manage withdrawal symptoms in newborns who have been exposed to opioids or other substances in utero. \\
    \hline
    \end{tabular}
    \label{tab:llm_error}
\end{center}
\end{table*}

\begin{table*}
\caption{Examples of mismatched conditions by the human expert (revised after criteria clarification)}
\begin{center}
    \begin{tabular}{|p{2.7cm}|p{3.5cm}|p{10.6cm}|}
        \hline
        \textbf{Patient record} & \textbf{Suggested mapping} & \textbf{Explanation} \\
        \hline
        Severe intraventricular haemorrhage &  Grade IV preterm intraventricular haemorrhage & Although the second description refers to a more specific condition, Grade IV implies that it is severe, hence the MD accepted this mapping.\\
        \hline
        Mechanical ventilation  &  Respiratory failure requiring assisted ventilation & The second description refers specifically to respiratory failure, whereas mechanical ventilation is a method of treatment. Since it is mentioned in the EHR, it implies the patient's respiratory failure, hence the MD accepted this mapping.\\
        \hline
        Adverse events (AEs)  &  Adverse drug reaction (ADR) & AEs are any undesirable medical occurrences, not necessarily caused by the drug. Since our dataset describes outcomes of drug use in pregnancy, the MD assumed the AE is an ADR.\\
        \hline
        5-minutes Apgar $<7$  &  5-minute APGAR score of 0 & The first description is more generic.\\
        \hline
    \end{tabular}
    \label{tab:human_error}
\end{center}
\end{table*}

To summarize, the mapping of this dataset would be a much harder task without the generated suggestions based on vector embeddings. The medical researchers did not know how to reliably map these data using conventional ontology-based search methods. 
The patient records for mapping are not easy to interpret: some of them are very short, others include abbreviations or ambiguous syntax. As we showed via our evaluation, the suggested mappings based on the vector space similarity alone is not good enough.

Based on this study, we conclude that validation of the mappings with an LLM significantly improves the mapping precision. We observed that the LLM had certain difficulties in deciding whether to accept mappings for similar but not identical records. It is important to formulate acceptance criteria for most basic ontological relations (such as the subtype- or class-inclusion-relation and the part-whole relation). 

\subsection{Annotated dataset}
The second dataset included information similar to that of the previous experiment, but the outcomes were already annotated with ICD-10 ontology codes. Hence, to align this dataset with the target ontologies, it was sufficient to translate 1162 unique ICD-10 codes featured in the dataset to the corresponding alternatives in MONDO and/or HPO.  

Although MONDO and HPO provision fields for cross-references to other ontologies, a quick search revealed that in the case of mapping from/to ICD-10, these references are available only for a small fraction of diseases, namely, 1840 for MONDO, and 39 for HPO. Considering the size of these ontologies (HPO currently contains over 13,000 terms and over 156,000 annotations to hereditary diseases, MONDO defines approximately 25,600 disease terms, and the ICD-10 classification allows for more than 14,000 different codes), it is not an exaggeration to say that the identifier-based mapping between them is not directly available.

\begin{figure}[htbp]
    \centering
    \includegraphics[width=0.48\textwidth]{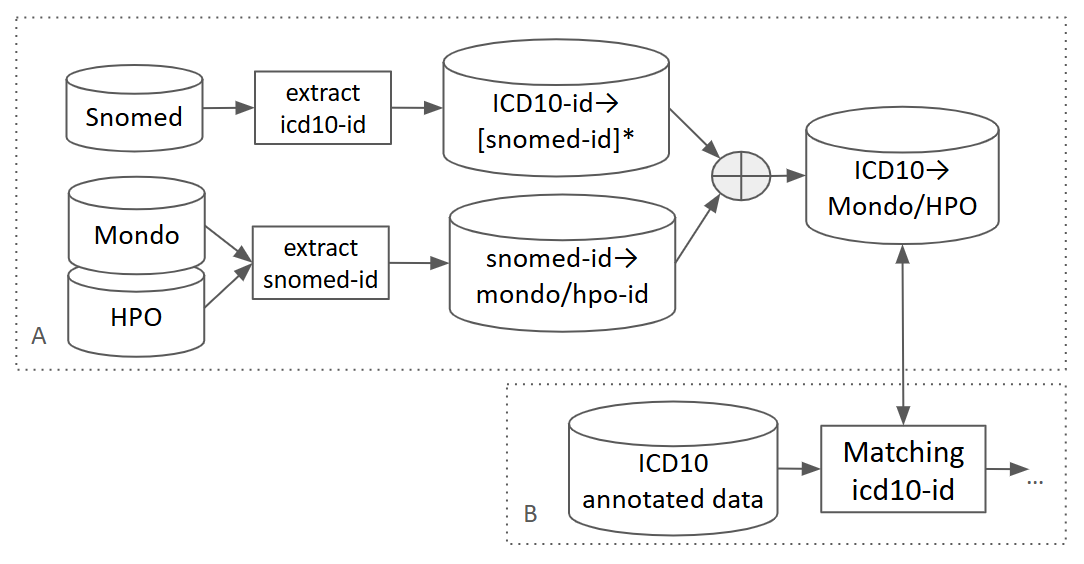}
    \caption{ICD-10-to-MONDO/HPO conversion via SNOMED, candidate pair generation}
    \label{fig:via-snomed}
\end{figure}

To bridge the annotations, we used two methods:
\begin{itemize}
    \item The RAG-based method relies on the embedded vector search, as in the previous example. Similarly, we searched for 3 best matches. This generator produced 3129 candidate pairs. 
    \item The SNOMED-based generator, outlined in Figure~\ref{fig:via-snomed}, connects the input ICD-10 codes to the target MONDO/HPO codes via the SNOMED CT database. This database provides references to ICD-10 while MONDO and HPO have cross-references with  SNOMED. 
    This method had no limit on how many mappings to produce for each ICD-10 code, we paired all ICD-10 codes with all MONDO/HPO codes related to the same SNOMED identifier. This generator produced 7787 candidate pairs.
\end{itemize}

It is important to emphasize that mapping SNOMED CT to ICD-10 presents several challenges due to differences in structure, granularity, and intended use between the two systems. SNOMED CT is a comprehensive clinical terminology designed for detailed patient records, while ICD-10 is a classification system primarily used for statistical and billing purposes.
The study by Wang and Bodenreider~\cite{wang2013synergism} concludes that a single SNOMED CT concept may require multiple ICD-10 codes to fully represent its meaning. Furthermore, the appropriate ICD-10 code can depend on patient-specific factors such as age and comorbidities, which require rule-based mapping approaches. Another study~\cite{miller2019using} emphasizes the practical difficulties in mapping and highlights the need for careful consideration of the clinical context.
Due to these reasons, not all pairs of ICD-10 and MONDO/HPO codes formed by the cross-reference search on SNOMED CT refer to the same disease, and validation via the acceptance prompt is still necessary. 

\begin{figure}[htbp]
    \centering
    \includegraphics[width=0.48\textwidth]{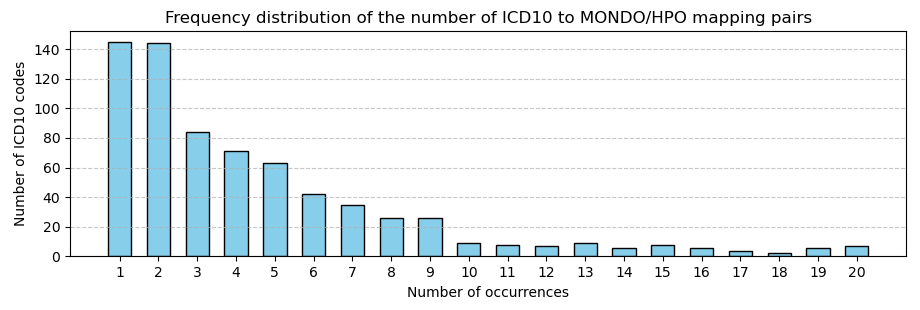}
    \caption{ICD-10-to-MONDO/HPO conversion via SNOMED, number of inputs vs target mappings}
    \label{fig:histogram}
\end{figure}

The mapping method outlined in Figure~\ref{fig:via-snomed} produced 7787 matching pairs involving 800 original ICD-10 codes. For 362 ICD-10 codes no results were retrieved, either because (i) ICD-10 was not mentioned in SNOMED CT (192 cases) or (ii) SNOMED CT code was not mentioned in MONDO and HPO references (170 cases). Figure~\ref{fig:histogram} shows part of the distribution of the number of relevant matches per input code retrieved via the SNOMED CT database. This distribution is extremely right-skewed; the image omits the entries that map into 20 or more codes. Although most of the ICD-10 codes (98\%) map to 10 MONDO and HPO terms or less, 2\% of ICD-10 conditions translate into 10 or more relevant MONDO and HPO terms, with extreme cases reaching over hundred of relevant mappings (see Table~\ref{tab:snomed-multi}). 

\begin{table}
\caption{Examples of ICD-10 codes with the large number of relevant MONDO and HPO terms}
\begin{center}
    \begin{tabular}{|p{1cm}|p{6cm}|p{0.5cm}|}
        \hline
        ICD-10 & ICD-10 label & n \\
        \hline
            Q878 & Other specified congenital malformation syndromes, not elsewhere classified & 329\\
            Q870 & Congenital malformation syndromes predominantly affecting facial appearance & 287\\
            Q828 & Other specified congenital malformations of skin & 195\\
            Q788 & Other specified osteochondrodysplasias & 143 \\
            Q872 & Congenital malformation syndromes predominantly involving limbs & 136\\
        \hline
    \end{tabular}
    \label{tab:snomed-multi}
\end{center}
\end{table}

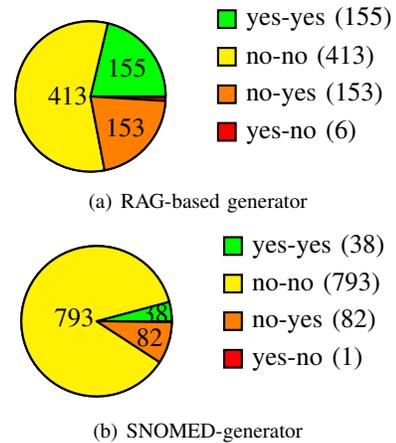
\begin{figure}[htbp]
    \centering
    \subfigure[RAG-based generator]{
    \label{fig:icd10-pie1}
    \begin{tikzpicture}
        \pie[
            text=legend,
            sum=728,   
            hide number,
            radius=1,    
            color={green, yellow, orange, red} 
        ]{
            155/yes-yes (155),
            413/no-no (413),
            153/no-yes (153),            
            6/yes-no (6)
        }
        \pie[sum=728, radius=1, color={green, yellow, orange}]{155/,413/,153/}
    \end{tikzpicture}
    }
    \subfigure[SNOMED-generator]{\begin{tikzpicture}
    \label{fig:icd10-pie2}
        \pie[
            text=legend, 
            sum=915,   
            hide number,
            radius=1,    
            color={green, yellow, orange, red} 
        ]{
            38/yes-yes (38),
            793/no-no (793),
            82/no-yes (82),            
            1/yes-no (1) 
        }     
        \pie[
            sum=915, radius=1, color={green, yellow, orange} 
        ]{
            38/,793/,82/
        }
    \end{tikzpicture}}
    \caption{Number of data record mappings approved and rejected by a human expert vs LLM. }
    \label{fig:ICD-10-eval}
\end{figure}

In both cases, LLM was making the final decision whether to accept or reject the mappings, accepting pairs with an equivalent or more generic output. In the case of embedding-based matching, 42.3\% of matching pairs were accepted. In the case of the SNOMED-based conversion, 14.7\% of the matching pairs were accepted. Figure~\ref{fig:ICD-10-eval} shows the precision assessment of human expert versus LLM for subsets (728 and 915 random records, respectively) of the records in both versions of ICD-10 to MONDO and HPO matching sets. In the RAG-generated pipeline, MD and LLM agree on 78\% of decisions. With SNOMED-based matching pair generation, MD and LLM agree on 91\% of the entries. The evaluation datasets and scripts to implement the presented mapping pipeline are available in~\cite{zenodo-dataset}. 


For more relaxed acceptance criteria, i.e., whether both descriptions refer to the same disease or to the related but more general or more specific condition, the acceptance ratios were 71\% and 80\%, respectively. Both RAG and SNOMED-based methods generated good candidate pairs, but the output condition was often more restrictive.    
Interestingly, the target datasets produced by two pair generation methods differed significantly: only for 475 ICD-10 codes (out of 1162) a MONDO or an HPO term generated by the RAG-based method was also among the codes produced by the SNOMED-based method. This can be explained by the following observations:
\begin{itemize}
    \item The recall of the RAG-based mapping pipeline may be compromised by our decision to retrieve only three relevant terms. As the alternative mapping method revealed, only 54\% of the ICD-10 codes were mapped to 3 or less terms via SNOMED. Retrieving 10 top matching pairs would ensure better recall by producing relevant options for 98\% of entries. However, this would significantly increase the workload on the human expert evaluating LLM's precision and result into a lower acceptance rate like in the case of the SNOMED-based generator. 
    \item The SNOMED-based mapping did not produce any suggestions for 31\% ICD-10 codes.  
\end{itemize}
Hence, to improve mapping recall, it is useful to combine candidate pairs from both generators and rely on the LLM to filter out mappings not suitable for a particular research question. 

\section{Related work}\label{sect:related-work}

Data harmonization in FL is complex due to varying formats, terminologies, and standards across decentralized data sources. In sensitive domains like pediatric care, this is further complicated by privacy and consent requirements, necessitating standardized frameworks and interoperable validation to enable compliant, collaborative research.

Schmidt et al.~\cite{schmidt2020harmonisation} survey published research papers to identify common definitions, goals, and workflows for data harmonization in healthcare. The review identified six common terms for data harmonization, including record linkage and health information exchange, and outlined nine key components such as integrating multiple databases, using unique patient identifiers, and involving data across various levels and institutions. The report concludes that data completeness, quality, and coding were common barriers to effective use in clinical decision-making. 

Nan et al.~\cite{nan2022fusion} provide a comprehensive review of data harmonization techniques in digital healthcare, highlighting methodological trends, challenges, and a proposed checklist to guide future fusion-based applications. The review focuses on the computational data harmonization approaches for multi-modal data. Rolland et al.~\cite{rolland2015harmonization} propose a structured, six-step process to harmonize cancer epidemiology data, aiming to improve the reproducibility and rigor of pooled multi-study analyses. 

The CVD-COVID-UK consortium developed a four-layer harmonization method using large-scale EHRs to enable efficient analysis of COVID-19 and cardiovascular diseases across the UK~\cite{abbasizanjani2023harmonising}. The method successfully harmonized data for over 59 million individuals, offering a transparent, scalable approach for multi-nation research, which has supported various studies, particularly on COVID-19's cardiovascular impact. Adhikari et al.~\cite{adhikari2020cohort} focuses on cohort studies, offering practical guidance for managing and harmonizing data to enable multi-study integration and improve statistical power.

Topaloglu and Palchuk present TriNetX\cite{Topaloglu2018}, a clinical research collaboration platform that enables data-driven study design without requiring centralized data pooling. This application highlights the growing need for secure and privacy-preserving federated data analysis. The deployment in a secure, standards-compliant virtual cloud environment further underscores the critical role of secure infrastructure in supporting federated research networks.

Swarm Learning~\cite{PMID:34040261} is a decentralized ML approach designed to enable the use of sensitive medical data across institutions without violating privacy laws. Unlike traditional FL, it uses edge computing and blockchain-based coordination without a central server. Results showed that Swarm Learning models outperformed local models while preserving data confidentiality, offering a promising path for privacy-preserving precision medicine.

Stonebraker and Ilyas~\cite{stonebraker2018data} review the evolution of data integration systems, highlighting the limitations of traditional approaches such as ETL (extract, transform, and load) pipelines and federated databases. They emphasize the growing importance of addressing semantic heterogeneity and the need for more automated, scalable ML-driven methods. The authors propose a shift toward declarative, AI-assisted solutions to manage the increasing complexity and volume of heterogeneous data sources.

Gibson et al.~\cite{Gibson2021} explore the feasibility of employing ML techniques to develop claims-based algorithms for identifying health outcomes of interest (HOIs), specifically focusing on rhabdomyolysis (a condition in which damaged skeletal muscle breaks down rapidly). The study demonstrated that ML models, particularly the Super Learner ensemble, achieved higher positive predictive values compared to traditional expert-developed models, indicating the potential of these techniques to enhance the accuracy and efficiency of electronic phenotyping in healthcare research.

Prisma~\cite{Hellenberg2025PRISMAAP} is a generic schema matching approach that leverages functional dependencies (FDs) to capture relationships between columns, even in cases where data is encrypted or column names are cryptic. Prisma uses a four-step process involving profiling databases, filtering FDs, creating graph-based representations, and comparing column embeddings to generate column correspondences. 

Several recent scientific studies have explored the application of LLMs in harmonizing healthcare data.
Fernandez et al.~\cite{10.14778/3611479.3611527} predicted that LLMs would disrupt data management in two key ways: by enabling semantic understanding to advance long-standing challenges such as entity resolution and schema matching, and by blurring the line between traditional databases and information retrieval systems. 

A. Santos et al.~\cite{santos2025harmonia} introduce a system that combines LLM-based reasoning with an interactive user interface and a library of data harmonization primitives. The system uses the top-k best matches between the source schema and the target schema. 
Matos et al.~\cite{matos2024ehrmonize} presents a framework that leverages LLMs to abstract medical concepts from EHRs. Evaluating five LLMs on tasks such as free-text extraction and binary classification, the research demonstrates that models like GPT-4o can achieve high accuracy in identifying generic route names and drug classifications, significantly enhancing efficiency in EHR data abstraction.

A study by Yikuan Li et al.~\cite{li2023fhir} investigates the capability of LLMs to enhance healthcare data interoperability by converting clinical texts into Fast Healthcare Interoperability Resources (FHIR) standards. The presented experiments demonstrate that LLMs can streamline natural language processing and achieve an exceptional accuracy rate in exact matches compared to human annotations. Arindam Sett et al.~\cite{sett2024standardizing} also explore the use of LLMs to standardize healthcare data by mapping clinical data schemata to established data standards like FHIR. The results indicate that the use of LLMs significantly reduces the need for manual data curation and improves the efficiency of the data standardization process, potentially accelerating the integration of AI in healthcare. Dukyong Yoon et al.~\cite{yoon2024interoperability} evaluated the performance of LLM in the transformation and transfer of healthcare data to support interoperability. Using MIMIC-III~\cite{johnson2016mimic} and UK Biobank datasets, the research demonstrates that LLMs can significantly improve data transformation and exchange, achieving high accuracy and efficiency without complex standardization processes.

Recent studies have explored the integration of LLMs like ChatGPT into healthcare~\cite{Wang2024,Abdullahi2024}. 
Wang et al.~\cite{Wang2024} screened 820 articles and included into the review 65 articles. Although LLMs have demonstrated potential in improving access to general medical information, medical knowledge retrieval, summarization, and administrative tasks, they are not always able to provide reliable answers to complex health-related tasks, e.g., diagnosis. Moreover, concerns persist regarding their reliability, biases, and privacy risks. 

Bhanbhro et at.~\cite{Bhanbhro2024} investigate FL challenges, focusing on data heterogeneity, client weighting, and resource disparities. Through experiments on datasets like MNIST, CIFAR-10, and brain MRI scans, the study demonstrates how non-IID (Independent and Identically Distributed) data distributions and varying client capabilities can adversely affect global model performance and convergence. The authors explore mitigation strategies such as weighted aggregation and model personalization, highlighting the trade-offs between data diversity, model accuracy, and system efficiency in FL environments.

Nasarian et al.~\cite{NASARIAN2024102412} review methods and challenges in implementing interpretable ML and explainable AI within healthcare. They propose a three-level interpretability process, preprocessing, modeling, and post-processing, to enhance clinician-AI communication and trust, offering a step-by-step roadmap for integrating responsible AI into clinical decision support systems.

Zhang et al.~\cite{Zhang2025} address ethical concerns in healthcare AI by introducing a resource-adaptive FL framework that promotes fairness and privacy. The proposed approach promises equitable participation across institutions with varying computational resources, improving model performance while protecting patient data.


Currently, the best-performing LLM models are commonly accessed via API requests. This practice raises concerns about data privacy, and organizations with strict data protection policies, such as healthcare centers, are hesitant to adopt LLM-based pipelines.    
This motivates the need for better open-source models which are competitive with closed-source models. Another solution is to use private LLMs instances and/or couple RAG and LLMs with differential privacy solutions~\cite{majmudar2022differentially}. 

While existing frameworks and studies have made significant progress in addressing specific challenges in data harmonization for FL -- ranging from schema alignment and semantic interoperability to privacy-preserving infrastructure and the integration of LLMs -- most have been tailored to particular domains, use cases, or workflows. In contrast, our work on the Brane/EPI frameworks proposes a more generalizable approach: a configurable programming environment designed to support a wide range of research workflows. By introducing a generic recipe for ontology-based data mapping and leveraging LLMs as semantic adjudicators, we aim to enable scalable, interpretable, and ontology-aligned federated research across heterogeneous datasets.

\section{Conclusions}\label{sect:conclusion} 
Human experts have the ability to analyze domain-specific and semantic content and perform data transformation. Often a team of researchers is involved (e.g., medical professional, ontology expert, data analyst, software developer), it is expensive and not reusable. LLM-assisted pipelines have a lot of potential in automating data alignment for federated research on private data that health organizations are not willing to disclose or do so only under rigid conditions within limited funded initiatives.  

A repository of ready-to-use data alignment functions that convert biomedical data between the most commonly accepted ontologies would be a valuable resource for any project that involves FL in healthcare. We showed that the two-step LLM-assisted conversion of data can be used effectively to align heterogeneous datasets to standardized vocabulary (the mapping precision in our experiments ranged from 78\% to 92\%), without or with very limited involvement of human expertise. 

Our future work aims at designing a low-code environment to foster FL within the Brane/EPI framework. This includes adoption of data-agnostic AI methods that can be integrated into scientific workflows as-is, model aggregation techniques, and LLM-based assistants to simplify the tasks performed by data scientists, such as workflow and policy definition or data alignment. 

\bibliographystyle{spmpsci}
\bibliography{main}
\end{document}